# Coordination and Control of Multiple Climbing Robots in Transport of Heavy Loads through Extreme Terrain

Himangshu Kalita[1], Steven Morad[2] and Jekan Thangavelautham[3]
*Space and Terrestrial Exploration (SpaceTREx) Laboratory, University of Arizona, 85721, USA*

**The discovery of ice deposits in the permanently shadowed craters of the lunar North and South Pole Moon presents an important opportunity for In-Situ Resource Utilization. These ice deposits maybe the source for sustaining a lunar base or for enabling an interplanetary refueling station. These ice deposits also preserve a unique record of the geology and environment of their hosts, both in terms of impact history and the supply of volatile compounds, and so are of immense scientific interest. To date, these ice deposits have been studied indirectly and by remote active radar, but they need to be analyzed in-situ by robotic systems that can study the depths of the deposits, their purity and composition. However, these shadowed craters never see sunlight and are one of the coldest places in the solar system. NASA JPL proposed use of solar reflectors mounted on crater rims to project sunlight into the crater depths for use by ground robots. The solar reflectors would heat the crater base and vehicles positioned at the base sufficiently to survive the cold-temperatures. Our approach analyzes part of the logistics of the approach, with teams of robots climbing up and down to the crater to access the ice deposits. The mission will require robots to climb down extreme environments and carry large structures, including instruments and communication devices.**

## I. Nomenclature

$g$ = acceleration due to gravity
$m_r$ = mass of each robot
$M$ = mass of payload
$I$ = moment of inertia of payload
$r_i$ = position vector of each robot
$v_i$ = velocity vector of each robot
$R$ = position vector of payload
$V$ = velocity vector of payload
$f_t$ = tension in tether
$f_{grip}$ = gripping force
$f_c$ = contact force
$f_f$ = frictional force

## II. Introduction

Beyond space exploration, the next big steps towards permanently living and working in space require developing critical space infrastructure to kick-start a space economy. Transporting fuel and material resources from Earth to space is expensive due to the 11 km/s escape delta-v. It is more cost-effective to use autonomous robots to mine and process resources from the neighboring Moon [27, 21] and Near-Earth Asteroids (NEAs) [26] for use in space, as they have much lower escape delta-vs. Water ice is a critical resource in this future economy as it can be processed into fuel for an interplanetary refueling station network. This could be through the use of water-steam [23], or electrolysis of water [24] to produce hydrogen and oxygen. Furthermore, hydrogen and oxygen maybe efficiently transported in the form of metal hydrides and perchlorates respectively [25]. They can release hydrogen and oxygen on demand to power high-energy Polymer Electrolyte Membrane (PEM) fuel cells [19, 28] with conversion efficiencies approaching

---

[1] PhD Student, Aerospace and Mechanical Engineering, University of Arizona.
[2] PhD Student, Aerospace and Mechanical Engineering, University of Arizona.
[3] Assistant Professor, Aerospace and Mechanical Engineering, University of Arizona.

55% to 70% in place of batteries to generate electricity. Batteries both primary and rechargeable have low specific energy that for the time being cannot match hydride-perchlorate based PEM fuel cell architectures [25]. These fuel cell system power systems may be used in eclipse to operate mining vehicles, transport shuttles [22], mobile devices and critical communication and navigation facilities [20] on the Moon and in cis-lunar space. The discovery of ice deposits in the permanently shadowed craters of the North and South Poles of Moon presents an important opportunity for In-Situ Resource Utilization. These ice deposits maybe the source for sustaining an interplanetary refueling station or even a lunar base. These ice deposits also preserve a unique record of the geology and environment of their hosts, both in terms of impact history and the supply of volatile compounds, and so are of immense scientific interest. To date, these ice deposits have been studied indirectly and by remote active radar, but they need to be analyzed in-situ by robotic systems that can study the depths of the deposits, their purity and composition. Fig. 1 shows the distribution of locations with potential water ice trapped in permanently shadowed regions of the Moon along with the maximum annual temperature distribution acquired by the Moon Mineralogy Mapper ($M^3$), Lunar Orbiter Laser Altimeter (LOLA), Lyman Alpha Mapping Project (LAMP) and the Diviner Lunar Radiometer Experiment data set [1]. It reveals that the southern polar region exhibits a greater concentration of ice deposits compared to the northern pole.

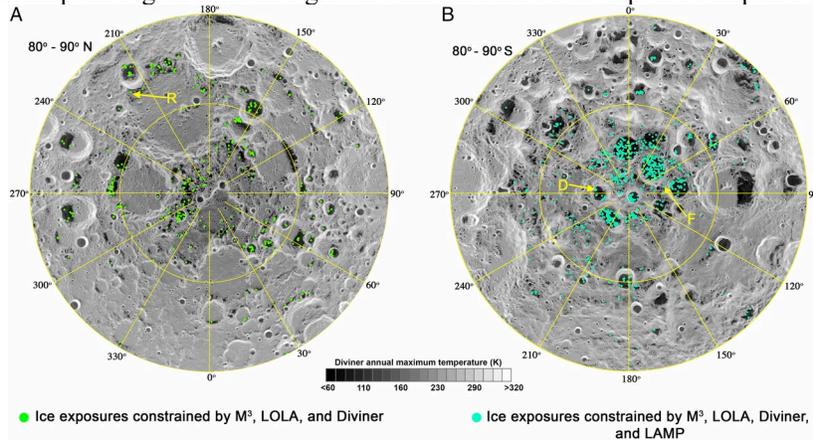

**Fig. 1:** Data points for presence of surface water ice in permanently shadowed regions at the North and South Polar regions of the Moon using $M^3$, LOLA, Diviner and LAMP data sets [1].

However, these shadowed craters never see sunlight and are one of the coldest places in the solar system. With solar reflectors mounted on the crater rims sunlight will be projected into the crater depths for use by ground robots [2]. The solar reflectors would heat the crater base and vehicles positioned at the base sufficiently to survive the cold-temperatures. With ground robots working in the crater base, there is a need for transportation of heavy loads like antennas, sloshing fuel tanks, rigid instruments from the crater rim to the base. In this work, we analyze the logistics of the ISRU activities, where a team of small, low-cost robots climb up/down the crater walls to transport heavy loads. The robots organize themselves into a reconfigurable multirobot system to perform this mission. In the following paper, section III-VI presents background and relation work, system dynamics, analysis of the gripping mechanism, and dynamics simulations respectively. Section VII presents a way to find the optimal configurations of the robots connected to the payload using evolutionary algorithms followed by path planning methods in section VIII.

### III. Background

Extreme environment mobility and exploration has been researched widely in the last two decades with a multitude of robotic systems being proposed but climbing in these sloped natural terrains is still a daunting challenge. Much of the work in this field has been done on developing tethered, legged and wheeled robotic systems. A few examples include Dante II [3] for exploring volcanoes, Teamed Robots for Exploration and Science on Steep Areas (TRESSA) [4], Axel [5], All-Terrain Hex-Limbed Extra-Terrestrial Explorer (ATHELETE) [6] for exploring cliff faces. Other examples that uses friction, suction cups, magnets and sticky adhesives to climb slopes include Legged Excursion Mechanical Utility Rover (LEMUR IIb) [7], Stickybot [8], Spinybot II [9] and Robots in Scansorial Environments (RiSE) [10]. NASA JPL has also developed an anchoring foot mechanism for sampling on the surface of near-Earth asteroids using microspines that can withstand forces greater than 100N on natural rock and has proposed to use it on the Asteroid Retrieval Mission (ARM) [11]. However, all these examples are single robot systems and are prone to single point failure. Multiple robots operating as a team offer significant benefits over a single robot enabling distributed command and control.

Our past work has shown climbing strategies with small, low-cost, reconfigurable multirobot systems. Each robot is a 30 cm sphere of mass 3kg covered in microspines for gripping onto rugged surfaces and several robot attaches using passive tethers [29]. The robots secure itself to a slope using the microspine gripping actuators, and one by one each robot moves upwards/downwards by hopping up/down the slope [12, 13]. We had also shown autonomous path-planning and navigation of the robotic system in our past work [14]. The motivation for our work is taken from proven methods by alpinists to climb mountains. The mountaineers use ice axes and crampons to grip on the surface and climb steep mountain slopes. The use of legs and hands provide multiple contact points to the sloped surface and even when each attempt to grip onto a higher location fails, the climber is still secure with his feet and one hand gripping tightly onto the slope. In this work, we extended this approach to having the robots self-organize and carry heavy/bulky external loads up/down a slope.

## IV. System Dynamics

The system consists of a payload to be transported by $N$ spherical robots considered as point masses. The robots are connected to the payload by passive tethers as shown in Fig. 2. The vectors $x, y, z$ denotes the principle axes of the inertial frame attached to the inclined plane and vectors $\bar{x}, \bar{y}, \bar{z}$ denotes the principle axes of the body fixed frame attached to the payload. We define $m_r$ as the mass of each robot, $M$ as the mass of the payload, $I$ as the payload's moment of inertia about the $\bar{z}$ axis, $CG$ as the center of gravity of the payload and $CM$ as the center of mass of the robots. We also define $r_i$ and $v_i$ as the position and velocity vector of each robot w.r.t. the inertial frame, $R, V$ as the position and velocity vector of the payload w.r.t. the inertial frame, $\phi$ as the orientation of the payload w.r.t. the $x$ axis of the inertial frame and $p_i$ as the position vector w.r.t. the body fixed frame of the points to which the tethers are connected to the payload. Moreover, the angle between $-y$ axis and the vector connecting $CM$ and $CG$ is denoted as $\vartheta$, and the plane on which the system is climbing is inclined at an angle $\theta$.

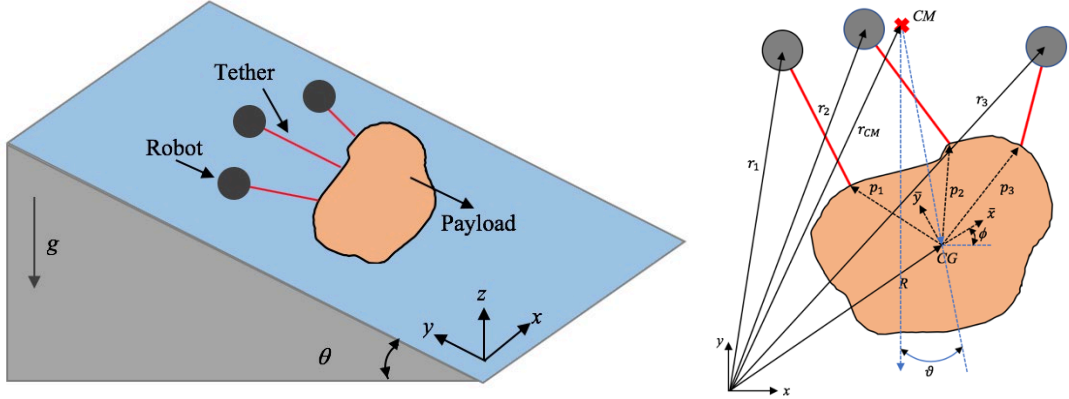

**Fig. 2: (Left) Schematic showing the multi-robot system climbing on an inclined slope carrying a heavy load. (Right) Schematic showing the reference frames and representing the holonomic and non-holonomic constraints between the robots and the payload.**

The equations of motion of the entire system can be written as:

$$\dot{r}_i = v_i, \quad \dot{R} = V \quad i = 1, \dots, N$$
$$m_r \dot{v}_i = f_{t(i)} + m_r f_g + f_c + f_{grip(i)} + u_{(i)}$$
$$M\dot{V} = -\sum_{i=1}^{N} f_{t(i)} + M f_g + f_c + f_f$$
$$I\ddot{\phi} = \sum_{i=1}^{N} (p_i \times f_{t(i)}) - k_r \phi - c_r \dot{\phi} \qquad (1)$$

where, $f_g = [0 \quad -g\sin\theta \quad -g\cos\theta]^T$ is the gravitational acceleration $f_{grip}$ is the gripping force which will be discussed in Section V and $u(i) = T/m_r$ is the control input with $T$ as the applied thrust. The tethers are modeled using Kelvin-Voight model as a viscoelastic material having the properties both of elasticity and viscosity. The tension in the tethers $f_t$ can be expressed as:

$$f_{t(i)} = \begin{cases} (k_t(\|l_i\| - l_0) + c_t v_{li})u_i & if \ \|l_i\| > l_0 \\ 0 & if \ \|l_i\| \le l_0 \end{cases} \qquad (2)$$

where $l_i = R + A^T p_i - r_i$, $u_i = l_i/\|l_i\|$, $v_{li} = (l_i^T \dot{l}_i)/\|l_i\|$, and $k_t$ and $c_t$ are the stiffness parameter and damping coefficient of the tether respectively. The contact force $f_c$ is determined by using the Hertz contact force model. Every collision consists of a compression phase and a restitution phase which can be modeled as a non-linear spring-damper as:

$$f_c = k_c \delta^n + c_c \dot{\delta} \quad (3)$$

where, $k_c$ is the stiffness parameter, which depends on the material properties and the local geometry of the contacting bodies, $\delta$ is the penetration depth, $c_c$ is the damping coefficient, $\dot{\delta}$ is the relative velocity of the contact points projected on an axis normal to the contact surfaces and $n = 3/2$. Also, $f_f$ is the viscous frictional force which is proportional to the relative velocity of the contacting surfaces and the coefficient of viscous friction $\mu_v$ and can be described in the simplest form as:

$$f_f = \mu_v V \quad (4)$$

## V. Gripping Mechanism

The performance of the multirobot system depends on the ability of each robot to grasp onto a rough natural surface. The gripping mechanism for each robot consists of an array of microspines embedded on a skin that wraps around its external surface. Simulations were done to understand the interaction of the spines with surfaces of different properties. A random surface generator is built to create 3D surfaces for a given RMS surface roughness value. Each spine is modeled as a curved beam with a circular cross section that tapers to a round tip of radius $r_s$. The spine approached the surface at an approach angle $\psi_a$ to engage to an asperity on the surface of radius $r_a$. The regions of the surface that a spine could grip on to is determined by computing the traced surface by a spine with tip radius $r_s$. Next the angle $\psi$ of the traced surface normal is determined and we search for the locations where $\psi$ is greater than $\psi_{min}$. The angle $\psi_{min}$ depends on the angle at which the spines are loaded $\psi_{load}$, and coefficient of friction $\mu$ between the spine and the surface [9].

$$\psi_{min} = \psi_{load} + \cot^{-1} \mu \quad (5)$$

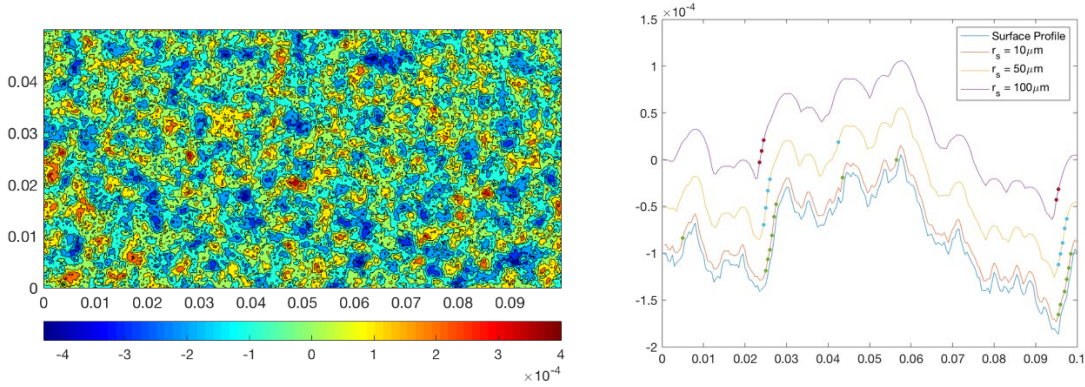

**Fig. 3: (Left) Contour map of surface with RMS surface roughness value of $100\mu m$. (Right) Surface traced by the center of spines of tip radius $r_s = 10\mu m, 50\mu m, 100\mu m$. The dots show the points where the respective spines can grip.**

Fig. 3 shows the profile of a 2D surface and the surface traced by the center of spines of different tip radius. It can be seen that smaller spines with smaller tip radius are more effective at engaging to asperities on smooth surfaces. The maximum load that a spine can sustain is a function of the tensile stress of the hook and square of the radius of curvature of the spine tip and asperity as follows:

$$f_{max} = \left[\left(\frac{\pi \sigma_{max}}{(1-2\mu)}\right)^3 \left(\frac{1}{2E^2}\right)\right] R^2 \quad (6)$$

$$\frac{1}{R} = \frac{1}{r_s} + \frac{1}{r_a} \quad (7)$$

Thus, as we decrease the tip radius of the spine, it can engage to smoother but the local load carrying capacity decreases. Hence, the design of the microspine skin has to be such that it can carry the load of the multirobot system and can engage onto a wide variety of rough surfaces. Each of the microspine toe consists an embedded elastic flexure mechanism that enables it to stretch parallel to the vertical surface under a load. Moreover, each spine can stretch and drag relative to its neighbors to find a suitable asperity to grip. If a toe catches an asperity, neighboring toes will continue to slide down as the caught toe stretches and grip on a suitable asperity. With $s$ number of spines attached to each robot, the maximum load each robot can carry is $f_{load} = \sum_{i=1}^{s} f_{max(i)}$. Thus, the gripping force $f_{grip}$ for each robot can be modeled as a sigmoid function as follows:

$$f_{grip(i)} = \frac{f_{load}}{1 + exp\left(-k\left(f_{t(i)} + m_r f_g - \frac{f_{load}}{2}\right)\right)} \tag{8}$$

where, $k$ defines the steepness of the curve and a value of $k = 15$ is used for the simulations.

## VI. Dynamics Simulations

Dynamics simulations were performed to validate the dynamics model and study the behavior of the system while climbing up and down a slope. Each robot needs to hop from its initial position $r_{0i}$ to its desired position $r_{\tau i}$. We can formulate an optimization problem to calculate the best hopping trajectory. The dynamics model discussed in section IV is used to solve the following optimization problem:

$$\min_{T,\tau} T_x^2 + T_y^2 + T_z^2$$

$$s.t. \begin{cases} \dot{r}_i = v_i \\ m_r \dot{v}_i = f_{ti} + m_r f_g + f_{grip} + \dfrac{T}{m_r} \\ r_i(0) = r_{0i} \\ r_i(\tau) = r_{\tau i} \\ \sqrt{T_x^2 + T_y^2 + T_z^2} \leq \|T\|_{max} \end{cases} \tag{9}$$

Where, $T = [T_x \ T_y \ T_z]^T$ is the thrust provided along $x, y, z$ axes respectively and $\|T\|_{max}$ is the maximum thrust each robot can provide. Dynamics simulations were carried out for the system to climb up a slope of $\theta = 15°$. The mass of each of the robots were $m_r = 1kg$. The payload was considered a circular disk of radius 1m, mass $M = 10kg$ and moment of inertia about the $\bar{z}$ axis as $I = 5kgm^2$. Fig. 4 shows the snapshots of the motion of the system, where the payload starts from its initial position $R_0 = [3 \ \ 2 \ \ 0]^T$ and is transported by 3 robots to its final position of $R_{final} = [3.1 \ \ 8.4 \ \ 0]^T$.

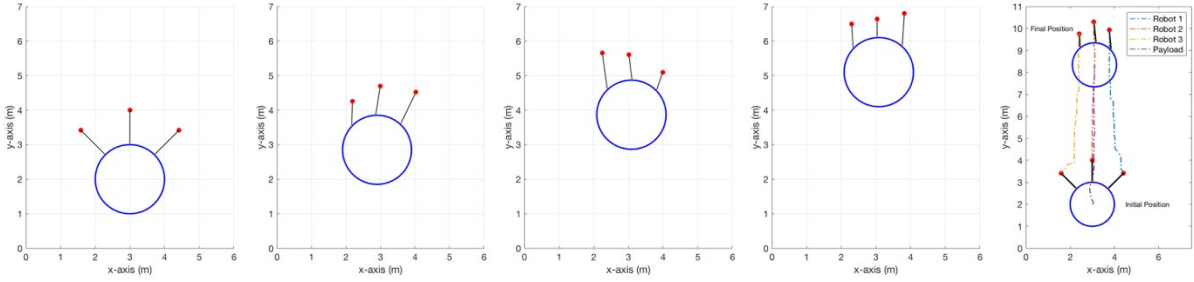

Fig. 4: Snapshots of the motion of the system while climbing up a slope of $\theta = 15°$.

Similar dynamics simulations were done for the system to climb down a slope of $\theta = 15°$. Fig. 5 shows the snapshots of the motion of the system, where the payload starts from its initial position $R_0 = [3 \ \ 8 \ \ 0]^T$ and is transported by 3 robots to its final position of $R_{final} = [2.1 \ \ 4.1 \ \ 0]^T$.

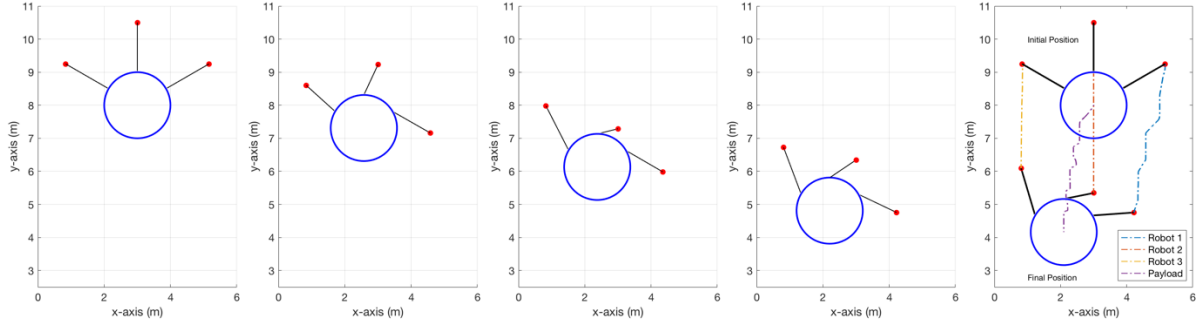

**Fig. 5: Snapshots of the motion of the system while climbing down a slope of $\theta = 15°$.**

Fig. 6 shows the $y$ position of each robot while climbing up and down a slope of $\theta = 15°$ which demonstrates the systematic climbing approach of the system with each robot hopping one at a time and gripping to the surface.

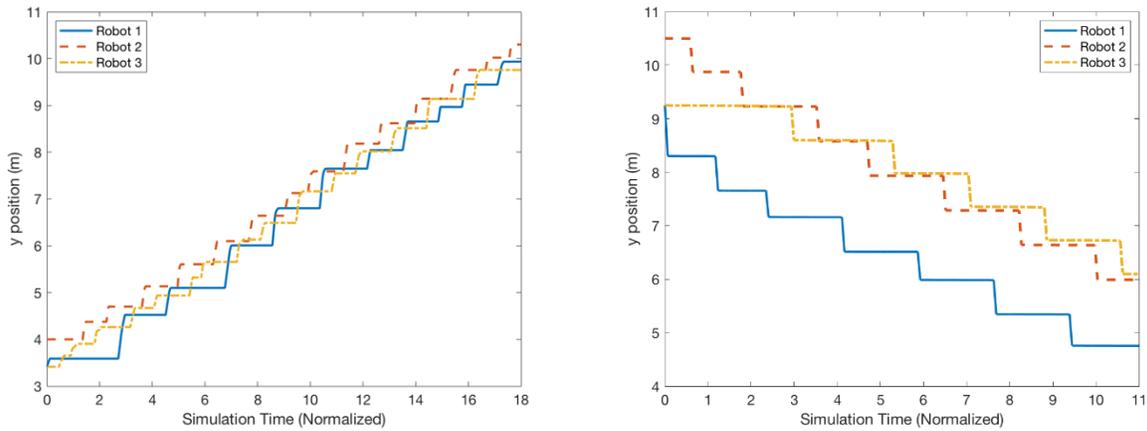

**Fig. 6: y-position of each robot showing systematic climbing approach. (Left) Climbing up and (Right) Climbing down a slope of $\theta = 15°$.**

## VII. Design Optimization Using Evolutionary Algorithm

The number of robots needed to carry a load and the positions at which they should be connected to the load will vary depending on the mass and inertia of the payload and climbing up or down a slope which is needed to be optimized. This is a scenario where Evolutionary Algorithms have potential, as they can generate good enough solutions through a directed, trial and error search. We used Evolutionary Algorithm to find the minimum number of robots and their positions to climb up/down a slope the maximum distance with minimum oscillations. The search space is defined by the angle $\alpha$ which divides the payload into $m$ nodes for robot connections and decides the size of the genotype which is equal to $m = 360/\alpha$. Each individual is represented by $m$ binary numbers so that they can be easily manipulated by standard genetic operators such as crossover and mutation. For each bit of the genotype, 1 represents a robot is connected to that node and 0 represents that no robot is connected to that node. Finally, the number of bits equal to 1 represents the number of robots connected to the payload. Fig. 7 shows the genotype structure for $\alpha = 15°$ and the equivalent robotic system.

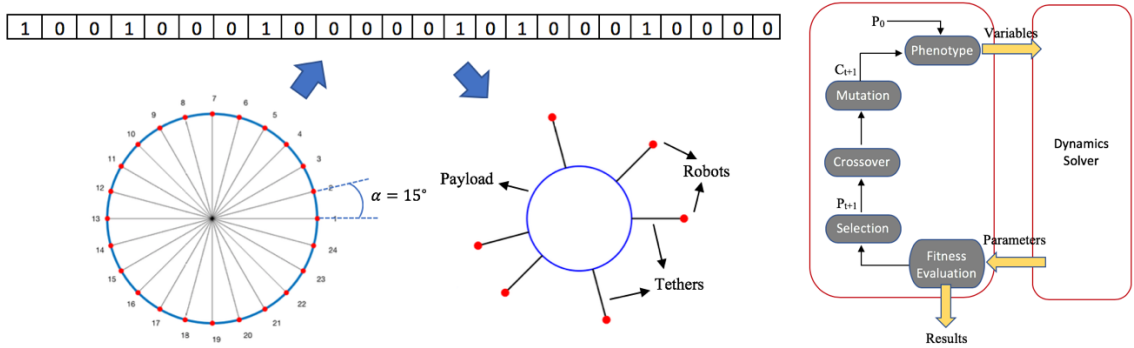

**Fig. 7:** (Left) Genotype structure showing the number of robots and their respective tether connection to the payload. (Right) Schematic of the optimization process using evolutionary algorithm.

The problem is formulated as a multi-objective optimization problem with three objectives. The first objective function is to maximize the distance travelled by the system at the expense of a given amount of energy (fuel). The second and third objectives are to minimize $\vartheta_{max}$ and $\overline{\vartheta}$ (mean value of $\vartheta$) in order to minimize the oscillation of the payload. A constraint is added to the problem so that no two tethers cross each other to avoid tangling of the tethers.

$$\begin{aligned} \max \quad & f_1 = \|R_{final} - R_0\| \\ \min \quad & f_2 = \vartheta_{max} \\ \min \quad & f_3 = \overline{\vartheta} \\ s.t. \quad & l_i \cap l_j = \emptyset, \quad i = 1, \dots, n, \ j = 1, \dots, n, \ i \neq j \end{aligned} \quad (10)$$

Each of the objectives are then normalized between 0 and 1 and then the overall fitness of the system is determined by taking the weighted sum as:

$$fitness = \alpha_1 f_1 + \alpha_2 f_2 + \alpha_3 f_3 \quad (11)$$

With the optimization problem defined, an elitist non-dominated sorting algorithm (NSGA-II) is used for the following work to find the pareto optimal solutions [15]. The initial parent population, $P_0$, is created of size $A$ which is then sorted based on the non-domination and then assigned a rank based on a cost function which is equal to its non-dominant level. The initial population then undergoes crossover and mutation to produce the set of offspring population $O_t$ of size $B$. Both the parents and children are then combined to produce $C_t = P_t \cup O_t$ of size $A + B$. The population $C_t$ is then sorted via non-dominance and assigned a rank. The first $A$ individuals of the set $C_t$ based on the non-dominant level is then selected for the next generation. The next population $P_{t+1}$ of size $A$ then again undergo selection, crossover and mutation. The process is repeated until the system achieves the desired fitness. For our analysis, we have considered an initial population of 50 with a crossover probability of 0.8 and a mutation probability of 0.2. For non-dominated sorting the 3 objective functions in (10) are used and finally a fitness is assigned to each individual based on a weighted sum as (11).

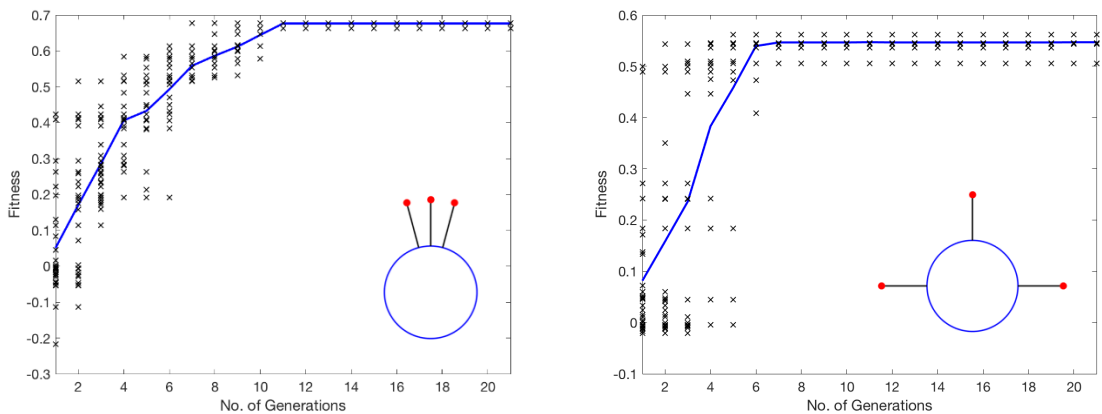

**Fig. 8:** Fitness of each individual and average fitness of each generation over 21 generations. (Left) Climbing up and (Right) Climbing down. The configuration shown on the bottom right corner are the fittest individual.

Fig. 8 shows the fitness of each individual and the average fitness of each generation over 21 generations. It was able to provide multiple solutions in the pareto front which can be analyzed further for final decision. The configurations shown on the bottom right corner of each figure shows the fittest individual evaluated by the optimization process. Further analysis can be done to find optimum configurations for different payload mass and inertia, slope, and surface roughness properties.

## VIII. Path Planning

We present an approach to tackle the problem of motion planning for multiple tethered robots. Due to the nature of multirobot off-world control, we aim to produce an online planning approach that runs quickly on the limited power hardware one might see on a rover. The main issue is performance, which arises from the high-dimensional space needed to represent the entire system. We make some simplifications to produce a tractable problem. A naive approach to motion planning with 3 robots and 1 payload would result in a 27-dimensional space: 18 dimensions for the robot poses, 6 dimensions for the pose of the payload, and 3 dimensions for the tether constraints. For each sample drawn by the planner, an integrator would need to compute the payload position, resulting in a problem that might take days to solve on space-grade computers. We instead make many simplifications to reduce the time and energy costs associated with motion planning. We reduce the problem to an inclined plane, removing pitch, roll, and $z$ position variables from each object, removing twelve dimensions. We assume the collision meshes of the hauling robots can be modeled as spheres, so we do not need to track yaw angle, removing two more dimensions. Next, we assume that the payload is heavy relative to the robots, and thus will follow the path of the robots in most cases.

A collision-free path for the robots in most cases provides a collision-free path for the payload as well. As such, we do not represent the position of the payload in the state space. Instead, a lazy planning algorithm validates the payload path is collision free once a path for the robots has been found. Checking for payload collisions only once a path is found greatly reduces the overhead of the payload position integration step. This heuristic allows the algorithm to run in real-time. However, this approach presents a problem because without the position of the payload, the robots will not stay together. To mitigate this concern, we can constrain the separation distance between robots. The triangle inequality is used to constrain the separation distance between each robot, keeping the robots together. The constraint is modeled as $\|pos0 - pos1\| < p/2$ and $\|pos1 - pos2\| < p/2$, where $p$ is the maximum allowed separation distance between all three robots.

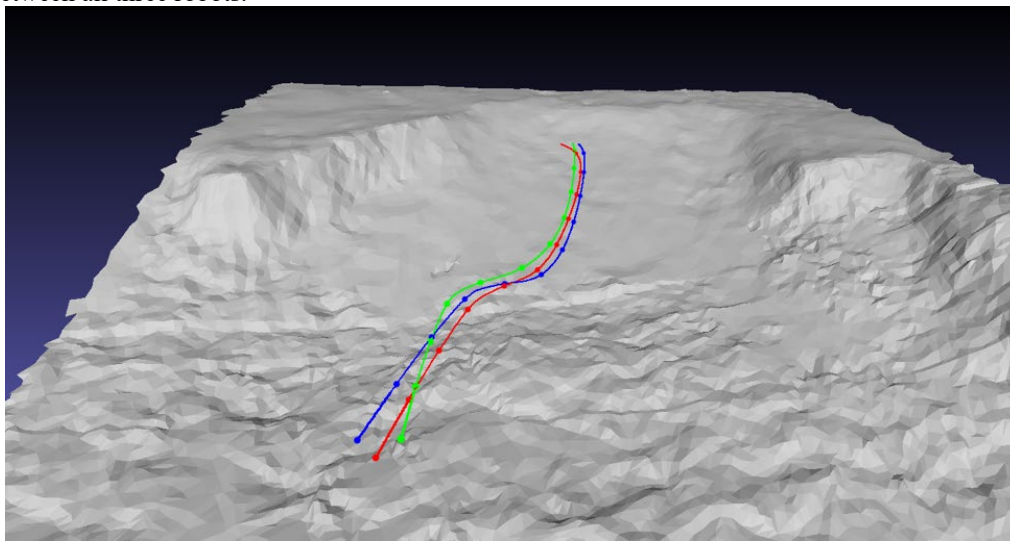

**Fig. 9: The LBKPIECE motion planner was run on a high-resolution mesh cutout of the Victoria crater on Mars. The red, green, and blue lines show the paths of each robot, with the dots denoting their hop landing points along the trajectory. Steep gradients were marked as obstacles. Note how the planner avoids the steep ridges to the left and right, and instead takes the relatively gentle slope up the middle. The planner took less than five seconds on a desktop computer to plan robot motion over the 256m$^2$ patch.**

This approach is implemented in C++ utilizing the OMPL library [16]. The six-dimensional state space is combined with the constraint to form a projection of the constrained state space in a lower dimensional space [17]. In this space, only states that satisfy the constraint exist. The Lazy Bi-directional Kinematic Planning by Interior-Exterior Cell Exploration [18] planner operates on this projected state space and produces a collision-free path that keeps the robots and payload safe. Our approach can plan motion over $256 m^2$ in less than five seconds using a desktop computer.

## IX. Conclusion

The paper presented a multirobot system for carrying heavy/bulky external load up/down a sloped rugged terrain. The system is prone to single-point failure and can withstand individual missteps, slips and falls during the climbing process. Dynamics and controls simulations of the system were presented which helped us understand the feasibility and behavior of the system. Evolutionary algorithms were used to analyze the robot configurations and maximize distance travelled and minimize oscillations of the hanging payload. A planning algorithm is also presented that was able to find trajectories for the system on crater walls while avoiding obstacles. The paper presented insights on the feasibility of the multirobot system in transporting heavy loads through extreme terrains in support of ISRU.